\documentclass[10pt, a4paper]{article}
\usepackage{lrec}
\usepackage{graphicx}
\usepackage{tabularx}
\usepackage{soul}
\usepackage[nolist]{acronym}

\usepackage{xspace}
\usepackage{todonotes}
\usepackage{url}
\usepackage{amssymb}
\usepackage{booktabs}
\usepackage{listings}
\usepackage{fancyvrb}
\usepackage{comment}
\usepackage{multirow} 
\usepackage{wrapfig}  
\usepackage[draft]{hyperref}
\usepackage{caption}

\usepackage{epstopdf}

\usepackage{hyperref}
\usepackage{xstring}
\usepackage{listings}
\definecolor{olivegreen}{rgb}{0.2,0.8,0.5}
\definecolor{grey}{rgb}{0.5,0.5,0.5}

\lstdefinelanguage{ttl}{
  basicstyle=\footnotesize \ttfamily,
  sensitive=true,
  morecomment=[l][\color{grey}]{@},
  morecomment=[l][\color{olivegreen}]{\#},
  morestring=[b][\color{blue}]\",
}

\lstdefinestyle{rdf}{numberblanklines=false, morekeywords={},
backgroundcolor=\color{backcolour},   
    commentstyle=\color{codegreen},
    keywordstyle=\color{magenta},
    stringstyle=\color{codeblue},
    basicstyle=\footnotesize,
    breakatwhitespace=false,         
    breaklines=true,                 
    captionpos=b,                    
    keepspaces=true,                 
    numbers=none,                    
    numbersep=5pt,                  
    showspaces=false,                
    showstringspaces=false,
    showtabs=false,                  
    tabsize=1
}

\lstdefinelanguage{SPARQL}{%
   morekeywords=[1]{CONSTRUCT,WHERE,SELECT},
   morekeywords=[2]{AND,FILTER,UNION,OPT,OPTIONAL,MINUS,ORDER,GROUP,BY,DESC,OFFSET,LIMIT},%
   morekeywords=[3]{sameTerm,isBLANK,isLITERAL,isIRI,BOUND,DISTINCT},
   morekeywords=[4]{rdf,rdfs,owl,dbo,res,xsd},
   morekeywords=[5]{>},
   morestring=[b]",%
   alsodigit={-},%
}[keywords,strings]

\newcommand{\triple}[1]{\texttt{(#1)}}
\newcommand{\resource}[1]{\texttt{#1}}

\newcommand{\literalPrefixed}[2]{\texttt{"#1"\textasciicircum\textasciicircum#2}}
\newcommand{\rdfLangString}{\texttt{rdf:langString}}

\usepackage{colortbl}
\usepackage{amsmath}
\usepackage{xcolor}
\usepackage{graphicx}
\usepackage{array}
\usepackage{makecell}
\usepackage{enumitem}

\newcommand{\rdfpt}{\textsc{RDF2PT}\xspace}

\begin{acronym}[UML]
	\acro{AOS}{Agricultural Ontology Services}
	\acro{AGRIS}{Agricultural Science and Technology}
	\acro{API}{Application Programming Interface}
	\acro{A2KB}{Annotation to Knowledge Base}
	\acro{BPSO}{Binary Particle-Swarm Optimization}
	\acro{BPMLOD}{Best Practices for Multilingual Linked Open Data}
	\acro{BFS}{Breadth-First-Search}
	\acro{CBD}{Concise Bounded Description}
	\acro{COG}{Content Oriented Guidelines}
	\acro{CSV}{Comma-Separated Values}
	\acro{CCR}{cross-document co-reference}
	\acro{DPSO}{Deterministic Particle-Swarm Optimization}
	\acro{DALY}{Disability Adjusted Life Year}
	\acro{D2KB}{Disambiguation to Knowledge Base}

	\acro{ER}{Entity Resolution}
	\acro{EM}{Expectation Maximization}
	\acro{EL}{Entity Linking}
	\acro{FAO}{Food and Agriculture Organization of the United Nations}
	\acro{GIS}{Geographic Information Systems}
	\acro{GHO}{Global Health Observatory}
	\acro{HDI}{Human Development Index}
	\acro{ICT}{Information and communication technologies}
    \acro{KB}{Knowledge Base}
	\acro{LR}  {Language Resource}
	\acro{LD}  {Linked Data}
	\acro{LLOD}  {Linguistic Linked Open Data}
	\acro{LIMES}{LInk discovery framework for MEtric Spaces}
	\acro{LS}  {Link Specifications}
	\acro{LDIF}{Linked Data Integration Framework}
	\acro{LGD} {LinkedGeoData}
	\acro{LOD} {Linked Open Data}
	\acro{MSE}{Mean Squared Error}
	\acro{MWE}{Multiword Expressions}
	\acro{NIF}{NLP Interchange Format}
	\acro{NIF4OGGD}{NLP Interchange Format for Open German Governmental Data}
	\acro{NLP}{Natural Language Processing}
	\acro{NER}{Named Entity Recognition}
	\acro{NED}{Named Entity Disambiguation}
	\acro{NEL}{Named Entity Linking}
	\acro{NN}{Neural Network}
	\acro{NLG}{Natural Language Generation}
	\acro{NN}{Neural Network}
	\acro{NL}{Natural Language}
	\acro{OSM}{OpenStreetMap}
	\acro{OWL}{Web Ontology Language}
	\acro{PFM}{Pseudo-F-Measures}
	\acro{PSO}{Particle-Swarm Optimization}
	\acro{QA}{Question Answering}
	\acro{RDF}{Resource Description Framework}
	\acro{SKOS}{Simple Knowledge Organization System}
	\acro{SPARQL}{SPARQL Protocol and RDF Query Language}
	\acro{SRL}{Statistical Relational Learning}
	\acro{SF}{surface forms}
	\acro{SW}{Semantic Web}
	\acro{UML}{Unified Modeling Language}
	\acro{WHO}{World Health Organization}
	\acro{WKT}{Well-Known Text}
	\acro{W3C}{World Wide Web Consortium}
	\acro{YPLL}{Years of Potential Life Lost}

	\acro{AOS}{Agricultural Ontology Services}
	\acro{AGRIS}{Agricultural Science and Technology}
	\acro{API}{Application Programming Interface}
	\acro{BPSO}{Binary Particle-Swarm Optimization}
	\acro{BPMLOD}{Best Practices for Multilingual Linked Open Data}
	\acro{CBD}{Concise Bounded Description}
	\acro{COG}{Content Oriented Guidelines}
	\acro{CSV}{Comma-Separated Values}
	\acro{CBMT}{Corpus-Based Machine Translation}
	\acro{CLIR}{Cross-Language Information Retrieval}
	\acro{DPSO}{Deterministic Particle-Swarm Optimization}
	\acro{DALY}{Disability Adjusted Life Year}

	\acro{ER}{Entity Resolution}
	\acro{EM}{Expectation Maximization}
	\acro{EBMT}{Example-Based Machine Translation}
	\acro{EBNF}{Extended Backus--Naur Form}
	\acro{EL}{Entity Linking}
	\acro{FAO}{Food and Agriculture Organization of the United Nations}
	\acro{GIS}{Geographic Information Systems}
	\acro{GHO}{Global Health Observatory}
	\acro{HDI}{Human Development Index}
	\acro{ICT}{Information and communication technologies}
    \acro{KB}{Knowledge Base}
	\acro{LR}  {Language Resource}
	\acro{LD}  {Linked Data}
	\acro{LLOD}  {Linguistic Linked Open Data}
	\acro{LIMES}{LInk discovery framework for MEtric Spaces}
	\acro{LS}  {Link Specifications}
	\acro{LDIF}{Linked Data Integration Framework}
	\acro{LGD} {LinkedGeoData}
	\acro{LOD} {Linked Open Data}
	\acro{MSE}{Mean Squared Error}
	\acro{MWE}{Multiword Expressions}
	\acro{MT}{Machine Translation}
	\acro{ML}{Machine Learning}
	\acro{NIF}{Natural Language Processing Interchange Format}
	\acro{NIF4OGGD}{NLP Interchange Format for Open German Governmental Data}
	\acro{NLP}{Natural Language Processing}
	\acro{NER}{Named Entity Recognition}
	\acro{NMT}{Neural Machine Translation}
	\acro{NN}{Neural Network}
	\acro{NLG}{Natural Language Generation}
	\acro{NED}{Named Entity Disambiguation}
	\acro{NERD}{Named Entity Recognition and Disambiguation}
	\acro{NL}{Natural Language}
	\acro{OSM}{OpenStreetMap}
	\acro{OWL}{Web Ontology Language}
	\acro{OOV}{out-of-vocabulary}
	\acro{PFM}{Pseudo-F-Measures}
	\acro{PSO}{Particle-Swarm Optimization}
	\acro{QA}{Question Answering}
	\acro{RDF}{Resource Description Framework}
	\acro{RBMT}{Rule-Based Machine Translation}
	\acro{SKOS}{Simple Knowledge Organization System}
	\acro{SPARQL}{SPARQL Protocol and RDF Query Language}
	\acro{SRL}{Statistical Relational Learning}
	\acro{SWT}{Semantic Web Technologies}
	\acro{SW}{Semantic Web}
	\acro{SMT}{Statistical Machine Translation}
	\acro{SWMT}{Semantic Web Machine Translation}

    \acro{TBMT} {Transfer-Based Machine Translation}
	\acro{UML}{Unified Modeling Language}
	\acro{WHO}{World Health Organization}
	\acro{WKT}{Well-Known Text}
	\acro{W3C}{World Wide Web Consortium}
	\acro{WSD}{Word Sense Disambiguation}
	\acro{YPLL}{Years of Potential Life Lost}

	\acro{AOS}{Agricultural Ontology Services}
	\acro{AGRIS}{Agricultural Science and Technology}
	\acro{API}{Application Programming Interface}
	\acro{A2KB}{Annotation to Knowledge Base}
	\acro{BPSO}{Binary Particle-Swarm Optimization}
	\acro{BPMLOD}{Best Practices for Multilingual Linked Open Data}
	\acro{BFS}{Breadth-First-Search}
	\acro{CBD}{Concise Bounded Description}
	\acro{COG}{Content Oriented Guidelines}
	\acro{CSV}{Comma-Separated Values}
	\acro{CCR}{cross-document co-reference}
	\acro{DPSO}{Deterministic Particle-Swarm Optimization}
	\acro{DALY}{Disability Adjusted Life Year}
	\acro{D2KB}{Disambiguation to Knowledge Base}

	\acro{ER}{Entity Resolution}
	\acro{EM}{Expectation Maximization}
	\acro{EL}{Entity Linking}
	\acro{FAO}{Food and Agriculture Organization of the United Nations}
	\acro{GIS}{Geographic Information Systems}
	\acro{GHO}{Global Health Observatory}
	\acro{HDI}{Human Development Index}
	\acro{ICT}{Information and communication technologies}
    \acro{KB}{Knowledge Base}
	\acro{LR}  {Language Resource}
	\acro{LD}  {Linked Data}
	\acro{LLOD}  {Linguistic Linked Open Data}
	\acro{LIMES}{LInk discovery framework for MEtric Spaces}
	\acro{LS}  {Link Specifications}
	\acro{LDIF}{Linked Data Integration Framework}
	\acro{LGD} {LinkedGeoData}
	\acro{LOD} {Linked Open Data}
	\acro{MSE}{Mean Squared Error}
	\acro{MWE}{Multiword Expressions}
	\acro{NIF}{NLP Interchange Format}
	\acro{NIF4OGGD}{NLP Interchange Format for Open German Governmental Data}
	\acro{NLP}{Natural Language Processing}
	\acro{NER}{Named Entity Recognition}
	\acro{NED}{Named Entity Disambiguation}
	\acro{NEL}{Named Entity Linking}
	\acro{NN}{Neural Network}
	\acro{OSM}{OpenStreetMap}
	\acro{OWL}{Web Ontology Language}
	\acro{PFM}{Pseudo-F-Measures}
	\acro{PSO}{Particle-Swarm Optimization}
	\acro{QA}{Question Answering}
	\acro{RDF}{Resource Description Framework}
	\acro{SKOS}{Simple Knowledge Organization System}
	\acro{SPARQL}{SPARQL Protocol and RDF Query Language}
	\acro{SRL}{Statistical Relational Learning}
	\acro{SF}{surface forms}
	\acro{UML}{Unified Modeling Language}
	\acro{WHO}{World Health Organization}
	\acro{WKT}{Well-Known Text}
	\acro{W3C}{World Wide Web Consortium}
	\acro{YPLL}{Years of Potential Life Lost}

 \acro{IRI}{Internationalized Resource Identifiers}
 \acro{URI}{Uniform Resource Identifier}
 \acro{RDF}{Resource Description Framework}
 \acro{OWL}{Web Ontology Language}
 \acro{MOS}{Manchester OWL Syntax}
 \acro{SPARQL}{SPARQL Query Language}
 
\end{acronym}

\title{RDF2PT: Generating Brazilian Portuguese Texts from RDF Data}

\name{Diego Moussallem\textsuperscript{1,2}, Thiago Castro Ferreira \textsuperscript{2}, Marcos Zampieri\textsuperscript{3}, \\ 
\textbf{\large{Maria Claudia Cavalcanti\textsuperscript{4}, Geraldo Xex\'eo\textsuperscript{5}, Mariana Neves\textsuperscript{6}, Axel-Cyrille Ngonga Ngomo\textsuperscript{1,7}}}}

 \address{%
     \textsuperscript{1}University of Leipzig, Germany,    
     \textsuperscript{2}Tilburg University, The Netherlands, \\
    \textsuperscript{3}University of Wolverhampton, United Kingdom, 
     \textsuperscript{4}Military Institute of Engineering - Brazil, \\
    \textsuperscript{5}Federal University of Rio de Janeiro - Brazil,
    \textsuperscript{6}German Federal Institute for Risk Assessment - Germany\\
    \textsuperscript{7}Paderborn University, Germany\\
    \textsuperscript{1}lastname@informatik.uni-leipzig.de\\
 }

\abstract{
The generation of natural language from \ac{RDF} data has recently gained significant attention due to the continuous growth of Linked Data. A number of these approaches generate natural language in languages other than English, however, no work has been proposed to generate Brazilian Portuguese texts out of \ac{RDF}. We address this research gap by presenting \rdfpt, an approach that verbalizes \ac{RDF} data to Brazilian Portuguese language. We evaluated \rdfpt in an open questionnaire with 44 native speakers divided into experts and non-experts. Our results suggest that \rdfpt is able to generate text which is similar to that generated by humans and can hence be easily understood. 
\\ \newline \Keywords{natural language generation, verbalization, semantic web}}

\begin{document}

\maketitleabstract

\section{Introduction}

\ac{NLG} is the process of generating coherent natural language text from non-linguistic data~\cite{reiter2000building}. Despite community agreement on the actual text and speech output of these systems, there is far less consensus on what the input should be~\cite{gatt2017survey}. A large number of inputs have been taken for \ac{NLG} systems, including images \cite{xu2015show}, numeric data~\cite{gkatzia2014comparing}, semantic representations \cite{theune2001data} and \ac{SW} data~\cite{ngonga2013sorry,bouayad2014natural}.

Presently, the generation of natural language from \ac{SW}, more precisely from \ac{RDF} data, has gained substantial attention~\cite{bouayad2014natural,staykova2014natural}. Some challenges have been proposed to investigate the quality of automatically generated texts from \ac{RDF}~\cite{colin2016webnlg}. Moreover, \ac{RDF} has demonstrated a promising ability to support the creation of \ac{NLG} benchmarks~\cite{gardent2017creating}. However, English is the only language which has been widely targeted. Even though there are studies which explore the generation of content in languages other than English, to the best of our knowledge, no work has been proposed to generate texts in Brazilian Portuguese from \ac{RDF} data. 

In this paper, we propose \rdfpt, a rule-based approach to verbalize \ac{RDF} data to Brazilian Portuguese. While the exciting avenue of using deep learning techniques in \ac{NLG} approaches~\cite{gatt2017survey} is open to this task and deep learning has already shown promising results for \ac{RDF} data~\cite{sleimi2016generating}, the morphological richness of Portuguese led us to develop a rule-based approach. This was to ensure that we could identify the challenges imposed by this language from the \ac{SW} perspective before applying \ac{ML} algorithms.

\rdfpt is able to generate either a single sentence or a summary of a given resource. 
In order to validate our approach, we evaluated \rdfpt using experts in \ac{NLP} and \ac{SW} as well as non-experts who are lay users or non-users of \ac{SW} technologies. Both groups are native speakers of Brazilian Portuguese.
The results suggest that \rdfpt generates texts which can be easily understood by humans and also help to identify some of the challenges related to the automatic generation of Brazilian Portuguese (especially from \ac{RDF}). The version of \rdfpt used in this paper, all experimental results and the texts generated for the experiments are publicly available.\footnote{\url{https://github.com/dice-group/RDF2PT}} 

 \section{Related Work}

According to~\newcite{staykova2014natural} and \newcite{bouayad2014natural}, there has been a plenty of works which investigated the generation of \ac{NL} texts from \ac{SWT} as an input data. 
However, the subject of research has only recently gained significant momentum. This attention comes from the great number of published works such as~\cite{cimiano2013exploiting,duma2013generating,ell2014language,biran2015discourse} which used \ac{RDF} as an input data and achieved promising results. Also, the works published in the WebNLG \cite{colin2016webnlg} challenge, which used deep learning techniques such as \cite{sleimi2016generating,mrabet2016aligning}, also contributed to this interest. \ac{RDF} has also been showing promising benefits to the generation of benchmarks for evaluating \ac{NLG} systems~\cite{gardent2017creating,perez2016building,mohammed2016category,schwitter2004controlled,hewlett2005effective,sun2006domain}.

Despite the plethora of works written on handling \ac{SWT} data, only a few have exploited the generation of languages other than English, for instance, ~\newcite{keet2017toward} to Zulu language. Additionally, a considerable number of \ac{NLG} approaches can be found to European or Brazilian Portuguese 
languages~\cite{pereira2008statistical,cuevas2008machine,de2009testbed,de2010text,de2012portuguese,de2013portuguese,de2014adapting,pereira2015towards}, however, none of them have exploited the generation of \ac{NL} from \ac{RDF}. Therefore, to the best of our knowledge, RDF2PT is the first proposed approach to this end.  

\section{The \rdfpt approach}
\label{sec:approach}

In this section, we first give a brief description of how \ac{RDF} data is able to represent useful linguistic information and we detail our approach \rdfpt in a sequence.

 \subsection{Preliminary RDF concepts}
 \label{subsec:rdf}

 Although previous works such as \newcite{sun2006domain} have already introduced how a single \ac{RDF} statement contains linguistic information, we briefly explain the concept for a better understanding of \rdfpt. 

 \ac{RDF}~\cite{rdf-concepts} statements are based on graph data models for representing knowledge. Thus, an RDF graph is a set of facts. Facts are expressed as so-called triples in the form \triple{subject predicate object}. The \emph{subjects} and \emph{predicates} are \ac{IRI}s 
and \emph{objects} are either \ac{IRI}s or literals. Literals, in general, have a datatype that defines its kind of values. For example, a literal can be a date, a number, a measure, a word or a group of words. On the other hand, a predicate denotes a binary relation between the \emph{subject} and \emph{object} as an argument. Additionally, \ac{RDF} vocabulary comprises some built-in properties. The most common one is \texttt{rdf:type}, which states that a resource denoted by the subject is an instance of the class specified by the object of the triple. For example, the \autoref{lst:example} shows a fragment of Albert Einstein DBpedia's resource\footnote{\url{http://dbpedia.org/resource/Albert_Einstein}} which represents the following information: \emph{``Albert Einstein was a scientist who worked in physics area. He was born in Ulm and died in Princeton."}.
 
\begin{lstlisting}[label=lst:example,language=ttl,caption=An excerpt of RDF triples.]
:Albert_Einstein rdf:type dbo:Scientist
:Albert_Einstein dbo:field :Physics
:Albert_Einstein dbo:birthPlace :Ulm 	
:Albert_Einstein dbo:deathPlace :Princeton 	
\end{lstlisting}

\subsection{Approach}

\rdfpt approach is akin to the approach SPARQL2NL~\cite{ngonga2013sorry} from which the project SemWeb2NL\footnote{\url{https://github.com/AKSW/SemWeb2NL}} originated. SemWeb2NL comprises rule-based and template-based approaches which aim to verbalize texts and concepts not only from \ac{RDF} triples but also from ontologies and SPARQL queries into English. 
In addition, SemWeb2NL is able to produce automatically educational \ac{QA} systems for self-assessment~\cite{buhmann2015assess}.
Despite the \rdfpt approach being capable of generating single sentences from distinct \ac{RDF} triples, for the sake of space, our description focuses on how \rdfpt can output a simplified summary of a given resource. 

A generic \ac{NLG} pipeline is composed by three tasks which are \emph{Document Planing}, \emph{Micro Planning} and \emph{Realization}. \rdfpt operates mostly at the level of the first two and to the \emph{Realization} task, \rdfpt uses an adaption of SimpleNLG to Brazilian Portuguese~\cite{de2014adapting}.


In the following sections, we describe the \rdfpt steps according to an \ac{NLG} system pipeline~\cite{gatt2017survey}. We then use the Portuguese version of DBpedia as a \ac{KB}~\cite{auer2007dbpedia,lehmann2015dbpedia} and as source for our examples.  

\subsection{Document Planning}
\label{subsection:document_planning}

This initial phase is divided into two sub-tasks. First, \emph{Content determination}, which is responsible for deciding what information a certain \ac{NLG} system should include in the generated text. Second, \emph{Discourse planning} (also known as Text structuring), which determines the order of the information in paragraphs and its rhetorical relation.

\paragraph{Content determination}
\label{subsubsection:content_determination}

\rdfpt assumes the description of a resource to be the set of \ac{RDF} statements of which this resource is the subject. Hence, given a resource, \rdfpt first performs a SPARQL query to get its most specific class through the predicate \texttt{rdf:type}. Afterward, \rdfpt gets all resources which belong to this specific class and ranks their predicates by using Page Rank~\cite{page1999pagerank} over the \ac{KB}. By these means, our approach can determine the most popular facts of this specific class.\footnote{The predicates can vary according to the classes and \ac{KB}}. Once the predicates are ranked, \rdfpt considers only the top seven most popular predicates of the class to describe the input resource.\footnote{This choice was based on~\newcite{gardent2017creating} which states that seven triples is a reasonable number for describing a resource.} For example, given \texttt{dbr:Albert\_Einstein} as a resource, \rdfpt determines its most specific class to be \texttt{dbo:Scientist}. Then, it ranks all the predicates used by this class per popularity according to its resources (see \autoref{lst:predicate}).
 
\begin{lstlisting}[label=lst:predicate,language=ttl, caption=Most popular predicates of a scientist.]
rdf:type                    dbo:field
dbo:deathPlace              dbo:almaMater
dbo:knownFor                dbo:award
dbo:doctoralStudent
\end{lstlisting}

\paragraph{Discourse planning}
\label{subsubsection:dicourse}

In this step, \rdfpt clusters and orders the triples. The subjects are ordered with respect to the number of their occurrences, thus assigning them to those input triples that mention them.
\rdfpt processes the input in descending order with respect to the frequency of the variables they contain, starting with the projection variables and only after that, turning to other variables. This method has already been used by other approaches and is the most effective method to follow regarding rule-based approaches to \ac{RDF}~\cite{bouayad2014natural}. As an example, consider the following triples in \autoref{lst:beforeordering}.

\begin{lstlisting}[label=lst:beforeordering,language=ttl,basicstyle=\scriptsize \ttfamily, caption=Example of triples before planning]
:Albert_Einstein  dbo:deathPlace :Princeton.
:Princeton        dbo:Country    :USA.
:Albert_Einstein  rdf:type       :Scientist.
:Albert_Einstein  dbo:knownFor   :General_relativity.
:Albert_Einstein  dbo:knownFor   :Brownian_motion.
:Albert_Einstein  dbo:birthPlace :Ulm.
:Ulm              rdf:type       :City.
:Ulm              dbo:Country    :Germany.
\end{lstlisting}

\autoref{lst:beforeordering} presents three subjects, \resource{:Albert\_Einstein}, \resource{:Ulm} and \resource{:Princeton}. As \resource{:Albert\_Einstein} is assigned to more triples than the others, it takes the first place in the discourse, followed by \resource{:Ulm},\resource{:Princeton} respectively (see \autoref{lst:afterordering}). However, \rdfpt still considers the popularity of predicates from the previous steps and organizes triples based on it, for instance, \texttt{rdf:type} comes before others due to its frequency in the \ac{KB}. 

\begin{lstlisting}[label=lst:afterordering,language=ttl,basicstyle=\scriptsize \ttfamily, caption=Example of triples after planning]
:Albert_Einstein  rdf:type       :Scientist
:Albert_Einstein  dbo:birthPlace :Ulm
:Albert_Einstein  dbo:deathPlace :Princeton
:Albert_Einstein  dbo:knownFor   :General_relativity.
:Albert_Einstein  dbo:knownFor   :Brownian_motion.
:Ulm              rdf:type       :City
:Ulm              dbo:Country    :Germany
:Princeton        dbo:Country    :USA
\end{lstlisting}

\subsection{Micro Planning}
\label{subsection:content determination}

This step is concerned with the planning of a sentence. It comprises three sub-tasks. Firstly, \emph{Sentence aggregation} decides whether information will be presented individually or separately. Second, \emph{Lexicalization} chooses the right words and phrases in natural language for expressing the semantics about the data. Third, \emph{Coreference generation} (also known as \emph{Referring expression}) is the task responsible for generating syntagms (references) to discourse entities, for example, whether the text should refer to an entity using a definite description, a pronoun or a proper noun \cite{ferreira2016towards}. In the following, we describe the challenges behind the tasks entailed.

\paragraph{Sentence aggregation}
\label{subsec:aggregation}
This task is based on~\newcite{ngonga2013sorry}. It 
is divided into two phases, \emph{subject grouping} and \emph{object grouping}. \emph{Subject grouping} collapses the predicates and objects of two triples if their subjects are the same. \emph{Object grouping} collapses the subjects of two triples if the predicates and objects of the triples are the same. 

The common elements are usually subject noun phrases and verb phrases (verbs together with object noun phrases). In order to maximize the grouping effects, we additionally collapse common prefixes and suffixes of triples, irrespective of whether they are full subject noun phrases or complete verb phrases.

In \autoref{lst:groupping}, the predicate \texttt{dbo:knownFor} shares the same subject \texttt{:Albert\_Einstein}  and also has two objects, \texttt{:General\_relativity} and \texttt{:Brownian\_motion}. Additionally, the predicate \texttt{dbo:birthPlace} shares  the same object \texttt{:Ulm} and has two subjects, \texttt{:Albert\_Einstein} and \texttt{:Gabriel\_Steiner}. They therefore can be collapsed using a conjunction \texttt{AND}, thus organizing and omitting repetitive triples. Moreover, we remove repetitions that arise when triples' verbalizations lead to the same natural language representation.  

\begin{lstlisting}[label=lst:groupping, caption=Grouping subjects and objects, language=ttl,basicstyle=\scriptsize \ttfamily]
 :Albert_Einstein dbo:knownFor :General_relativity.
 :Albert_Einstein dbo:knownFor :Brownian_motion.
 :Albert_Einstein dbo:birthPlace :Ulm.
 :Gabriel_Steiner dbo:birthPlace :Ulm.
\end{lstlisting}

\paragraph{Lexicalization} 
\label{pagraph:lexi}
This step comprises the main contribution of \rdfpt for verbalizing the triples in Brazilian Portuguese. 
In contrast to English, Brazilian Portuguese is a morphologically rich language which contains the grammatical gender of words. Grammatical gender plays a key role because it affects the generation of determiners and pronouns. It also influences the inflection of nouns and verbs. For instance, the passive expression of the verb \texttt{nascer} (en: ``be born") is \texttt{nascida} if the subject is feminine or \texttt{nascido} if masculine. Thus, the gender of words is essential for comprehending the semantics of a given Portuguese text. Also, Brazilian Portuguese has different possibilities in the expression of subject possessives. 
Hence, \rdfpt has to deal with the following phenomena while lexicalizing: 

\begin{itemize}

\item \textbf{Grammatical gender} - In Portuguese, the gender varies between masculine and feminine. This variation leads to supplementary challenges when lexicalizing words automatically. For example, a gender may be represented by articles ``um" and ``o" (masculine) or ``uma" and ``a" (feminine). However, the gender also affects the inflection of words. For instance, for the word ``cantor" (en: ``singer"), if the subject is feminine, the word becomes ``cantora". However, there are words which do not inflect, e.g., the word ``gerente" (en: ``manager"). If the subject is a woman, we only refer to it by using the article ``a'', i.e., ``a gerente". Therefore, there are some challenges to tackle for recognizing the gender and assigning it correctly. A tricky example to solve automatically is ``O Rio de Janeiro \'{e} uma cidade" (en: Rio de Janeiro is a city). In this case, the subject is masculine but its complement is feminine. Devising handcrafted rules to handle these phenomena can become a hard task. To address this challenge, we use a Part-Of-Speech tagger (TreeTagger in our case) as it retrieves the gender along with the parts of speech.\footnote{see the POS tags \url{http://www.cis.uni-muenchen.de/~schmid/tools/TreeTagger/data/Portuguese-Tagset.html}}
All the obtained genders are attached along with the lexicalizations for supporting the realization step. 

\item  \textbf{Classes and resources} - The lexicalizaton of classes and resources is carried by using a SPARQL query to get their Portuguese labels through the \texttt{rdfs:label} predicate\footnote{Note that it could be any property which returns a natural language representation of the given URI, see \cite{ell2011}.}. In case such a label does not exist, we use either the fragment of their URI (the string after the \verb|#| character) if it exists, or the string after the last occurrence of ``\verb|/|". Finally, this natural language representation is lexicalized as a noun phrase. Afterwards, \rdfpt recognizes the gender. In case the resource is recognized as a person, \rdfpt applies a string similarity measure (0.8 threshold) between the lexicalized word with a list of names provided by SemWeb2NL. This list is divided by masculine and feminine which in turn results in the gender. If the resource is not a person, we use Tree-tagger.

\item \textbf{Properties} - The lexicalization of properties relies on one of the results of~\newcite{ngonga2013sorry}, i.e., that most property labels are either nouns or verbs. To determine which lexicalization to use automatically, we rely on the insight that the first and last words of a property label in Portuguese are commonly the key to determining the type of property. We then use the Tree-Tagger to get the part of speech of predicates. Properties whose label begins with a verb are lexicalized as verbs. For example, the predicate \texttt{dbo:knownFor}, which Portuguese label is ``conhecido por", has the first word identified as an inflection of the verb ``conhecer" (en:know). Therefore, \rdfpt lexicalizes and sets it as a verb.  We devised a set of rules to capture this behavior, which we omit due to space restrictions.\footnote{All rules can be found in our code.} Moreover, \rdfpt uses some pre-defined templates for improving the quality of lexicalization. For example, the predicate \texttt{dbo:birthPlace}, \rdfpt uses the verb ``nascer" (eng: be born) along with the predicate ``em" (en: ``in"), so this predicate can be lexicalized as ``nasceu em" (en: was born in).  

For predicates which are recognized as nouns, \rdfpt relies on labels. For instance, \texttt{dbo:birth\-Date} is labeled as ``data de nascimento" and recognized as a noun phrase because of its first word ``data". \rdfpt also uses the first word of predicates to set the gender. For example, \texttt{dbo:deathPlace} is transliterated as ``local de falecimento". ``local" is masculine. Hence, the determiner to be used in front of this predicate needs to be ``o". In contrast to \texttt{dbo:birthDate} (``data de nascimento"), the word ``data" is feminine, thus the determiner must be "a".

\item \textbf{Literals} - In an RDF graph, literals usually consist of a \emph{lexical form} \texttt{LF} and a \emph{datatype IRI} \texttt{DT}. If the datatype is \rdfLangString, a non-empty \emph{language tag} is specified and the literal is denoted as a \emph{language-tagged string}.\footnote{In RDF 1.0 literals have been divided into ``plain' literals with no type and optional language tags, and typed literals.}
Accordingly, the lexicalization of strings with language tags is carried by using simply the lexical form, while omitting the language tag. For example, 
\texttt{``Albert Einstein"@pt} is lexicalized as ``Albert Einstein" or 
\texttt{"Alemanha"@pt} (``Germany"@en) is lexicalized as ``Alemanha". For other types of literals, we  differentiate between built-in\footnote{List of data types:\url{http://tinyurl.com/y95mxyxa}} and user-defined datatypes. 
For built-in literals, we use the lexical form, e.g., \literalPrefixed{123}{xsd:int} $\Rightarrow$ ``123". User-defined types are processed by using the literal value together with the (pluralized) natural language representation of the datatype IRI. Thus, we lexicalize \literalPrefixed{123}{dt:squareKilometre} as ``123 quil\^{o}metros quadrados" (en: ``123 square kilometres").

\end{itemize}


\paragraph{Coreference generation}
\label{sec:coreference}

In this step, \rdfpt relies on the number of subjects contained by the \ac{RDF} statements and only uses other expressions to refer to a given subject in case there is more than one mention of it. \rdfpt replaces the subject by possessive or personal pronouns with the corresponding gender depending on the predicates. For instance, given a triple \texttt{dbr: Albert\_Einstein dbo:birthPlace dbr:Ulm}, the predicate is a noun phrase then the subject is replaced by a possessive form which is ``seu" (en:``his"). However, Brazilian Portuguese has two different ways to express possession and this variation exists due to the necessity of handling complex syntaxes in some sentences and also because the gender of pronouns agrees with objects instead of subjects. For example, ``A professora proibiu que o aluno utilizasse \texttt{seu} dicion\'ario." (eng: ``The teacher forbade the student to use \texttt{his/her} dictionary''). The possessive pronoun \texttt{seu} in this sentence does not indicate explicitly to whom the dictionary belongs, if it belongs to the \texttt{professora} (eng:teacher) or \texttt{aluno} (eng:student). Thus, we have explicitly to define the possessive pronoun in order to decrease the ambiguity in texts and it is obviously important when generating text from data. If this sentence was translated into English, we would have indicated to whom the dictionary belonged, \texttt{her} or \texttt{his}. To this end, we handle the ambiguity of possessive pronouns by interspersing the alternative forms, e.g., \texttt{dele} (eng:his) or \texttt{dela} (eng: her)" which agrees with the subject. However, it is used just in case more than one subject exists in the same description.

In case the predicate is recognized as a verb (e.g, \texttt{dbr: Albert\_Einstein dbo:knownFor dbr:General\_relativity}), the subject is replaced by its respective personal pronoun \texttt{ele} (eng: ``he"). While setting the pronouns, \rdfpt recognizes the gender's subject. The \texttt{dbo:knownFor} is a verb phrase, thus the subject is replaced by the personal pronoun ``:ele"(see \autoref{tab:coreference}).

\begin{table}[htb]
\footnotesize
\centering
\begin{tabular}{@{} l @{}}
\toprule
\textbf{Triples before co-reference} \\
\toprule
1 - (Albert Einstein, ser, cientista) \\
2 - (Albert Einstein, local de nascimento, Ulm) \\
3 - (Albert Einstein, ser conhecido por, teoria da relatividade.) \\
\toprule
\textbf{Triples after co-reference} \\
\toprule
1 - (Albert Einstein, ser, cientista)\\
2 - (\textbf{seu}, local de nascimento, Ulm.)\\
3 - (\textbf{ele}, ser conhecido por, teoria da relatividade).\\
 \bottomrule
\end{tabular}
\caption{Example of triples in the coreference generation task.}
\label{tab:coreference}
\end{table}
	
\subsection{Linguistic realisation}
\label{subsec:realisation}

This last step is responsible for mapping the obtained descriptions of sentences from the aforementioned tasks and verbalizing them syntactically, morphologically and orthographically into a correct natural language text. To this end, we perform this step by relying on a Brazilian adaptation of SimpleNLG~\cite{de2014adapting}.\footnote{See the complete list of dependency parsing tags in \newcite{ngonga2013sorry}.}

The realization of a triple \triple{s p o} depends mostly on the lexicalization of its predicate \texttt{p}. If \texttt{p} can be realized as a noun phrase, then a possessive clause can be used to express the semantics of \triple{s p o}. For example, if (\texttt{p}) is a relational noun like \texttt{death place} e.g. in the triple \triple{:Albert\_Einstein :deathPlace :Princeton}, then the verbalization is \texttt{o local de falecimento de Albert Einstein foi Princeton.} (eng: The death place of Albert Einstein was Princeton) which formally can be expressed as in equation~\ref{nounphrase}. In case there is a previous triple which shares the same subject, it would be \texttt{seu local de falecimento foi Princeton} (eng: his death place was Princeton).  

\begin{align}
    \rho(s, p, o) &\Rightarrow \text{\texttt{poss}}(\rho(p),\rho(s))\land \text{\texttt{subj}}\nonumber \\ & 
    (\text{\texttt{BE}},\rho(p))\land \text{\texttt{dobj}}(\text{\texttt{BE}},\rho(o))
    \label{nounphrase}
\end{align}

In case \texttt{p}'s lexicalization is a verb, then the triple is verbalized setting the predicate as a verb. For example, in \triple{:Albert\_Einstein :influenced :Nathan\_Rosen} $\rho(\texttt{p})$ is the verb \texttt{influenciar} (en: \texttt{influence}), thus, the verbalization is \texttt{Albert Einstein influ{\^e}nciou Nathan Rosen} (eng: Albert Einstein influenced Nathan Rosen) and may be formalized as in equation~\ref{verb}.

\begin{align}
    \rho(s, p, o) &\Rightarrow \text{\texttt{subj}}(\text{\texttt{BE}},\rho(p))\land \text{\texttt{dobj}}(\text{\texttt{BE}},\rho(o))
    \label{verb}
\end{align}

\rdfpt is able to merge sentences that were derived from the same cluster for generating a readable summary, thus resulting in coordinate sentences. For example, for the triples \triple{:Albert\_Einstein :birthPlace :Ulm} and \triple{:Albert\_Einstein :deathPlace :Princeton}, if $p_1$ and $p_2$ can be verbalized as nouns, then we apply the following rule:
\begin{align}
\rho(s, p_1, o_1)\land\rho(s, p_2, o_2) \Rightarrow \nonumber \\
\text{\texttt{conj}}(\text{\texttt{poss}}(\rho(p_1),\rho(s)) \nonumber \\ 
 \land \text{\texttt{subj}}(\text{\texttt{BE}}_1,\rho(p_1)) \land \text{\texttt{dobj}} (\text{\texttt{BE}}_1,\rho(o_1)) \land \text{\texttt{poss}}(\rho(p_2), \nonumber \\
\rho(\text{\texttt{pronoun}}(s))) \land \text{\texttt{subj}}(\text{\texttt{BE}}_2,\rho(p_2)) \nonumber\\
\land \text{\texttt{dobj}}(\text{\texttt{BE}}_2,\rho(o_2))
\end{align}

Note that \texttt{pronoun(s)} returns the correct pronoun for a resource based on its type and gender (see \autoref{subsection:content determination}). Therewith, we can generate \texttt{O local de nascimento de Albert Einstein é Ulm e seu local de falecimento {\'e} Princeton.} (eng: Albert Einstein's birthplace is Ulm and his death place is Princeton). In addition, in case the \ac{KB} provides the ending date of a given resource through some predicate, for example dbo:deathDate, \rdfpt is able to lexicalize all the verbs in the past tense.  

\section{Evaluation}

We based our evaluation methodology on~\newcite{gardent2017creating} and \newcite{ferreira2016towards}. Our main goal was to evaluate how well \rdfpt represents the information obtained from the data. We hence divided our evaluation set into expert and non-expert users. Both sets were made up of native speakers of Brazilian Portuguese. We selected six DBpedia categories like~\newcite{gardent2017creating} for selecting the topic of texts. The categories were Astronaut, Scientist, Building, WrittenWork, City, and University. We detail below how we carried out both evaluation sets.

\textbf{Experts} -  We aimed to evaluate the adequacy and fluency of the generated texts from the perspective of experts. All experts hold at least a master degree in the fields \ac{NLP} or \ac{SW}. In the questionnaire, we used the same two questions as \newcite{gardent2017creating}: (1) Adequacy: Does the text contain only and all the information from the data? (2) Fluency: Does the text sound fluent and natural? We asked the 10 experts to evaluate 12 texts distributed across the aforementioned DBpedia categories, with two pieces of text from each category. All texts were generated automatically by the \rdfpt approach. The answers were on a scale from 1 to 5.\footnote{Questionnaire:~\url{http://tinyurl.com/y9vegl4g}} 

\textbf{Non-experts} - We evaluated the clarity and fluency of the generated texts. To this end, we created three types of texts. First, the texts were generated using a baseline of \rdfpt approach, which removes the functional words and also does not apply coreference rules. This version served as baseline as there is no other work pertaining to generating Brazilian Portuguese from RDF. Second, we used the texts generated using the \rdfpt approach outline at \autoref{sec:approach} The third type of texts were created manually by three different human annotators. \autoref{tab:exampleText} depicts an example of text in the three versions.

\begin{table*}[htb]
\footnotesize
\centering
\begin{tabular}{@{} ll @{}}
\toprule
\textbf{Version} & \textbf{Text} \\
\toprule
Baseline & Albert Einstein \'e cientista, Albert Einstein campo \'e f\'isica, Albert Einstein lugar falecimento Princeton. Albert
Einstein \\
& ex-institui\c{c}\~ao \'e Universidade Zurique, Albert Einstein \'e conhecido Equivalência massa-energia, Albert Einstein pr\^{e}mio \'e \\
& Medalha Max Planck, Albert Einstein estudante doutorado \'e Ernst Gabor Straus.\\
\midrule
Modelo & Albert Einstein foi um cientista, o campo dele foi a f\'isica e ele faleceu no Princeton. Al\'em disso, sua ex-institui\c{c}\~ao foi \\ 
&a Universidade de Zurique, ele \'e conhecido pela Equival\^encia massa-energia, o pr\^emio dele foi a Medalha Max Planck e\\ 
&o estudante de doutorado dele foi o Ernst Gabor Straus.\\
\midrule
Humano & Albert Einstein era um cientista, que trabalhava na \'area de F\'isica. Era conhecido pela f\'ormula de equival\^encia entre massa \\ 
&e energia. Formou-se na Universidade de Zurique. Einstein ganhou a medalha Max Planck por seu trabalho. Em Princeton,\\
&onde morreu, teve sob sua orienta\c{c}\~ao Ernst Gabor Straus. \\
 \bottomrule
\end{tabular}
\caption{Example of text in the Baseline, \rdfpt approach and Human version.}
\label{tab:exampleText}
\end{table*}

In total, we created three versions of 18 texts (one text per resource) selected randomly from the aforementioned DBpedia categories (total: 54 texts). These texts were distributed over three lists, such that each list contained one variant of each text, and there was an equal number of texts from the three types (Baseline, \rdfpt, Human). The experiment was run on CrowdFlower and is publicly available.\footnote{\url{https://ilk.uvt.nl/~tcastrof/semPT/evaluation/}}

The experiment was performed by 30 participants (10 per list). They were asked to rate each text considering the clarity and fluency based on two questions from~\newcite{ferreira2016towards} on a scale from 1
(Very Bad) to 5 (Very Good). The questions were: (1) Fluency: Does the text present a consistent, logical flow? (2) Clarity: Is the text easy to understand?

\subsection{Results}
\label{sec:results}

\begin{figure}[htb]
\centering
\includegraphics[scale=0.25]{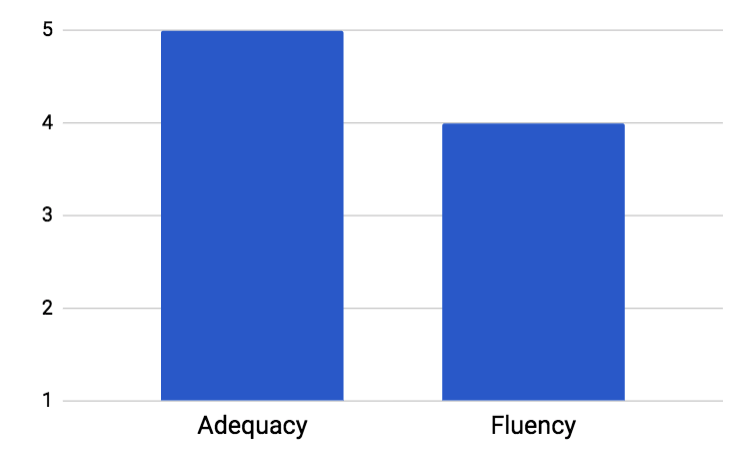}
\caption{\rdfpt results in experts survey}
\label{fig:experts}
\end{figure}

\textbf{Experts}~\autoref{fig:experts} displays the average fluency and clarity of the texts. The results suggest that \rdfpt is able to capture and represent the information from data adequately. Also, the generated texts are fluent enough to be understood by humans.

\begin{figure}[htb]
\centering
\includegraphics[scale=0.25]{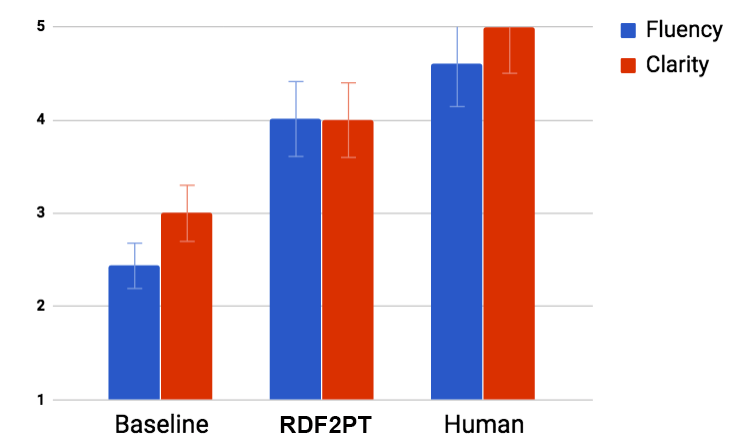}
\caption{Results in non-experts experiment}
\label{fig:non-experts}
\end{figure}

\textbf{Non-experts}~\autoref{fig:non-experts} depicts the average fluency and clarity of the texts where their topics are described by \emph{Baseline}, \emph{\rdfpt} and \emph{Human} approaches respectively. Inspection of this figure clearly shows that \emph{Baseline} texts are rated lower than both the \emph{\rdfpt} and \emph{Human} texts, in fact, \emph{\rdfpt} is superior to \emph{Baseline} and close to \emph{Human}.

We performed a statistical analysis in order to measure the significance of the difference between the types (Baseline, \rdfpt, Human). First, we carried out a Friedman test~\cite{friedman1937use} which resulted in a significant difference in the fluency ($\textit{x}^2$ = 193.61, $\rho$  \textless 0.0001) and clarity ($\textit{x}^2$ = 180.9, $\rho$  \textless 0.0001) for the three kinds of texts. Afterward, we conducted a post-hoc analysis with Wilcoxon signed-rank test corrected for multiple comparisons using the Bonferroni method, resulting in a significance level set at $\rho$  \textless 0.017. Texts of the Baseline are hence significantly less statistically understandable (Z=525 and $\rho$  \textless 0.017.) and fluent (Z=275.5 and $\rho$  \textless 0.017.) than those generated by the \rdfpt approach. However, \rdfpt also generates texts less comprehensible (Z=1617.5 and $\rho$  \textless 0.017.) and fluent (Z=1640.0 and $\rho$  \textless 0.017.) than those generated by humans. Clearly, humans were superior to Baseline in terms of comprehensibility (Z=234.5 and $\rho$  \textless 0.017.) and fluency (Z=264.0.0 and $\rho$  \textless 0.017.), as we expected. 
Therefore, there is a significant difference among all models, being baseline~\textless~model~\textless~human.      

\subsection{Discussion}
\label{para:discussion}

During the development of \rdfpt, some challenges to our rule-based algorithm became clear. The first challenge was to identify when the object of a predicate is an adjective. Consider the following triple, \triple{:Albert\_Einstein dbo:nationality :\'Austria}, its object \texttt{\'Austria} is a demonym and should be lexicalized as an adjective. However, it is lexicalized as a noun because the part-of-speech recognized by \rdfpt considers only the label \texttt{Austria}, which is a noun, and does not consider the predicate nationality, which is an important part, thus decreasing the quality of the generated texts. 
Second, \rdfpt's algorithm is totally dependent on ontology terms, thus when a given ontology contains wrong labels, \rdfpt is not able to recognize by itself the error and lexicalizes the terms wrongly. 
Third, the gender continues to be a hard task and \rdfpt sometimes presents poor results. For example, ``\texttt{Os Lus\'{i}adas \'e uns obra liter\'{a}ria}", the determiner \texttt{uns} should be feminine and singular, because \texttt{obra} is singular and has a feminine gender. However, it is accorded to the subject \texttt{Os Lus\'{i}adas}. This example is similar to the example presented in \autoref{pagraph:lexi} We hence envision the use of \ac{ML} algorithms for improving the gender recognition and generation. The last challenge observed was the generation of coordinated sentences by \rdfpt which helped the users in our experimental setup recognize if a given text was generated by \rdfpt or humans. This behavior is because  humans are likely to write subordinate sentences. For example, while \rdfpt is able to generate \texttt{Albert Einstein foi um cientista e ele nasceu em Ulm.} (eng: Albert Einstein was a scientist and he was born in Ulm), a human would write this same sentence in the following way, \texttt{Albert Einstein foi um cientista que/cujo nasceu em Ulm} (eng: Albert Einstein was a scientist who was born in Ulm). This difference was crucial in the perspective of our evaluators. Therefore, the generation of subordinate sentences in Portuguese must be investigated in the near future. 

\section{Further Application Scenarios}

We envision two promising applications using \rdfpt. The first aims to support the automatic creation of benchmarking datasets to \ac{NER} and \ac{EL} tasks. In Brazilian Portuguese, there is a lack of gold standards datasets for these tasks, which makes the investigation of these problems difficult for the scientific community. Our aim is to create Brazilian Portuguese silver standard datasets which are able to be uploaded into GERBIL\cite{gerbil} for an easy evaluation. To this end, we aim to implement \rdfpt in BENGAL~\cite{bengal}, which is an approach for automatically generating \ac{NER} benchmarks based on \ac{RDF} triples and Knowledge Graphs. This application has already resulted in promising datasets which we have used to investigate the capability of multilingual entity linking systems\footnote{\url{http://gerbil.aksw.org/gerbil/experiment?id=201801050040} and \url{http://gerbil.aksw.org/gerbil/experiment?id=201801110012}} for recognizing and disambiguating entities in Brazilian Portuguese texts. The second appealing application of \rdfpt is the generation of automatic QA systems based on \ac{RDF} for self-assessment. Therefore, the aim is to develop a Portuguese version of ASSES~\cite{buhmann2015assess}, which is a self-assessment platform for students based on DBpedia. 

\section{Conclusion}
We presented the \rdfpt approach which verbalizes \ac{RDF} data to Brazilian Portuguese texts. The results demonstrated that \rdfpt generates texts with a good quality of fluency and clarity compared to human texts. In addition, we identified important challenges for generating Brazilian Portuguese texts from \ac{RDF} using a rule-based approach. We intend to exploit the application of \ac{ML} models along with other Brazilian Portuguese resources\footnote{Resources: http://www.nilc.icmc.usp.br/nilc/index.php/tools-and-resources} to produce more fluent results and also to investigate the usage of \ac{ML} classification algorithms to improve the choice of grammatical gender of words.

Moreover, we aim to create a multilingual version of \rdfpt which will consist of French, German, Italian and Spanish. To this end, we will exploit the similarity among their syntaxes in the micro-planning task and we will reuse their respective SimpleNLG versions~\cite{mazzei2016simplenlg,bollmann2011adapting,vaudry2013adapting,aramossoto2017adapting} for the realization task.

\section*{Acknowledgments} 	

This work has been supported by the H2020 project HOBBIT (GA no. 688227)
and supported by the Brazilian National Council for Scientific and Technological Development (CNPq) (no. 206971/2014-1 and no. 203065/2014-0.). This research has also been supported by the German Federal Ministry of Transport and Digital Infrastructure (BMVI) in the projects LIMBO (no. 19F2029I), OPAL (no. 19F2028A) and GEISER (no. 01MD16014E) as well as by the BMBF project SOLIDE (no. 13N14456).


\section*{Bibliographical References}
\label{main:ref}

\bibliographystyle{lrec}
\bibliography{xample}


\end{document}